\documentclass{article} 
\usepackage[preprint]{colm2026_conference}

\usepackage{microtype}
\usepackage{graphicx}
\usepackage{subcaption}
\usepackage{booktabs} 
\usepackage{hyperref}
\usepackage{url}
\usepackage{lineno}

\definecolor{darkblue}{rgb}{0, 0, 0.5}
\hypersetup{colorlinks=true, citecolor=darkblue, linkcolor=darkblue, urlcolor=darkblue}

\usepackage{amsmath}
\usepackage{amssymb}
\usepackage{mathtools}
\usepackage{amsthm}

\usepackage[capitalize,noabbrev]{cleveref}

\theoremstyle{plain}

\theoremstyle{definition}

\theoremstyle{remark}

\usepackage[most]{tcolorbox}
\usepackage{xcolor}
\usepackage{colortbl}
\usepackage{enumitem}
\usepackage{wrapfig}
\usepackage{multicol}
\usepackage{tikz}
\usetikzlibrary{shapes.geometric, arrows.meta, positioning, fit, calc, decorations.pathreplacing}

\definecolor{userbg}{RGB}{255,230,220}
\definecolor{agentbg}{RGB}{220,240,255}
\definecolor{systembg}{RGB}{255,235,235}
\definecolor{goldenbg}{RGB}{255,248,220}
\definecolor{simbg}{RGB}{245,245,245}
\definecolor{goodgreen}{RGB}{50,160,50}
\definecolor{badred}{RGB}{200,50,50}
\definecolor{headerblue}{RGB}{60,120,180}
\definecolor{headerorange}{RGB}{220,120,50}
\definecolor{headerpink}{RGB}{200,80,120}
\definecolor{headergreen}{RGB}{60,150,80}
\definecolor{headerpurple}{RGB}{120,70,160}
\definecolor{warningbg}{RGB}{255,235,235}
\definecolor{prefbg}{RGB}{235,240,255}
\definecolor{exploreclr}{RGB}{70,130,200}
\definecolor{distillclr}{RGB}{200,120,50}
\definecolor{evolveclr}{RGB}{60,160,80}
\definecolor{inferclr}{RGB}{160,60,140}

\newcommand{\goodmark}{\textcolor{goodgreen}{\checkmark}}
\newcommand{\badmark}{\textcolor{badred}{$\boldsymbol{\times}$}}

\newcommand{\agenticon}[1]{%
\noindent\begin{minipage}[c]{0.08\linewidth}\centering\includegraphics[width=0.85\linewidth]{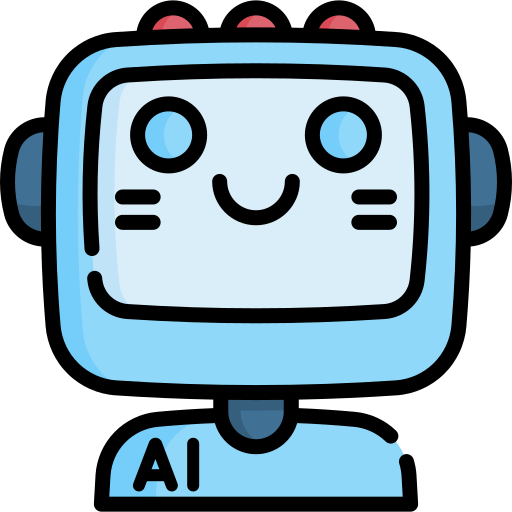}\end{minipage}%
\hfill%
\begin{minipage}[c]{0.89\linewidth}%
\begin{tcolorbox}[colback=agentbg, boxrule=0pt, arc=3pt, left=2pt, right=2pt, top=1pt, bottom=1pt, boxsep=1pt]%
\scriptsize \textbf{Agent:} #1%
\end{tcolorbox}%
\end{minipage}}
\newcommand{\usericon}[1]{%
\noindent\begin{minipage}[c]{0.89\linewidth}%
\begin{tcolorbox}[colback=userbg, boxrule=0pt, arc=3pt, left=2pt, right=2pt, top=1pt, bottom=1pt, boxsep=1pt]%
\scriptsize \textbf{User:} #1%
\end{tcolorbox}%
\end{minipage}%
\hfill%
\begin{minipage}[c]{0.08\linewidth}\centering\includegraphics[width=0.85\linewidth]{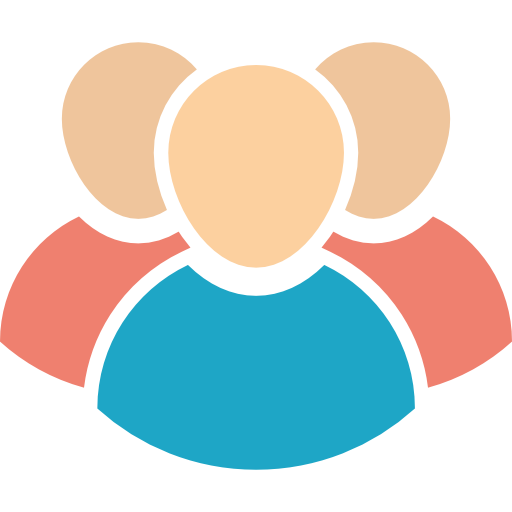}\end{minipage}}

\usepackage[disable,textsize=tiny]{todonotes}

\title{PRIME: Training Free Proactive Reasoning via Iterative Memory Evolution for User-Centric Agent}

\author{Prince Zizhuang Wang\\
  Carnegie Mellon University \\
  \texttt{princewang@cmu.edu} \\\And
  Shuli Jiang \\
  Carnegie Mellon University\\
  \texttt{shulij@alumni.cmu.edu} \\}

\begin{document}

\ifcolmsubmission
\linenumbers
\fi

\maketitle

\begin{abstract}
  The development of autonomous tool-use agents for complex, long-horizon tasks in collaboration with human users has become the frontier of agentic research. During multi-turn Human-AI interactions, the dynamic and uncertain nature of user demands poses a significant challenge; agents must not only invoke tools but also iteratively refine their understanding of user intent through effective communication. While recent advances in reinforcement learning offer a path to more capable tool-use agents, existing approaches require expensive training costs and struggle with turn-level credit assignment across extended interaction horizons. To this end, we introduce \textit{PRIME (Proactive Reasoning via Iterative Memory Evolution)}, a gradient-free learning framework that enables continuous agent evolvement through explicit experience accumulation rather than expensive parameter optimization. \textit{PRIME} distills multi-turn interaction trajectories into structured, human-readable experiences organized across three semantic zones: successful strategies, failure patterns, and user preferences. These experiences evolve through meta-level operations and guide future agent behavior via retrieval-augmented generation. Our experiments across several diverse user-centric environments demonstrate that \textit{PRIME} achieves competitive performance with gradient-based methods while offering cost-efficiency and interpretability. Together, \textit{PRIME} presents a practical paradigm for building proactive, collaborative agents that learn from Human-AI interaction without the computational burden of gradient-based training.
\end{abstract}

\section{Introduction}

Large Language Models (LLMs) \cite{achiam2023gpt,hui2024qwen2, team2023gemini, liu2024deepseek} are increasingly being deployed as autonomous agents capable of using tools to interact with the world. A primary challenge in this new paradigm is mastering dynamic, multi-turn dialogues with human users, which requires a dual capability: the technical skill to invoke tools and the communicative skill to understand a user's evolving intent~\cite{yao2024tau, barres2025tau}. Although Supervised Fine-Tuning (SFT) on expert data provides a baseline for agent behavior, it often results in rigid policies that fail to generalize to the unpredictable nature of real-world conversations. Training proactive agents that understand user intents while handling complex multi-turn tool-using tasks remains an open challenge~\cite{qian2025userbench, lu2024proactive, wu2025collabllm}.

Recent advances in reinforcement learning (RL) offer a compelling alternative, promising more robust and adaptive agents through interactive learning~\cite{team2025kimi}.
RL~\cite{schulman2017proximal, kool2019buy, ahmadian2024back, chen2025reinforcement, liu2025understanding, yu2025dapo} has proven effective for enhancing LLM capabilities in reasoning tasks, and the focus has recently shifted toward agents that interact with complex external tools and environments. Frameworks like \textit{ToolRL}~\cite{qian2025toolrl} and \textit{Sweet-RL}~\cite{zhou2025sweet} employ RL to teach agents to use external tools such as API functions for complex reasoning tasks. For user-centric tasks requiring proactive interaction with human users, \textit{UserRL}~\cite{qian2025userrl} and \textit{UserVille}~\cite{sun2025training} introduce multi-turn credit assignment to address sparse, delayed rewards in extended dialogues. These efforts have successfully equipped agents to operate tools within their respective environments, establishing a foundation for more complex interactive tasks.

However, gradient-based RL presents significant challenges for user-centric agents: multi-GPU training infrastructure is expensive, learned improvements are opaque and difficult to audit, and adaptation to new scenarios requires complete retraining cycles. To this end, we propose \textit{\textbf{PRIME} (\textbf{P}roactive \textbf{R}easoning via \textbf{I}terative \textbf{M}emory \textbf{E}volution)}, which takes a fundamentally different approach by treating agent improvement as a knowledge accumulation problem rather than a parameter optimization problem.
The central insight is that multi-turn user-centric interactions generate rich learning signals that can be captured, organized, and reused without gradient computation. When an agent successfully helps a user discover their hidden intent through clarifying questions, that successful strategy can be distilled into an explicit memory and retrieved when similar situations arise. When an agent fails by asking redundant questions that frustrate the user, that failure pattern can be documented and avoided in the future. This explicit knowledge representation enables continuous adaptation during deployment, transfer of learned strategies across different model architectures, and human inspection of what the agent has learned.
\textit{PRIME} achieves this through four interconnected phases that form a continuous learning cycle: \textit{exploration} of interaction environments to collect diverse trajectories, \textit{distillation} of those trajectories into structured memories with multi-turn credit assignment, \textit{evolution} of the memory library through meta-level optimization, and \textit{inference} where retrieved memories augment agent responses to better assist human users.
Our main contributions are as follows:
\begin{itemize}[leftmargin=*,itemsep=2pt,topsep=2pt]
    \item We introduce \textit{PRIME}, a gradient-free learning framework that improves user-centric agents through explicit memory accumulation rather than parameter optimization, eliminating the need for multi-GPU training while enabling interpretable, transferable improvements.
    \item We propose a memory distillation pipeline that leverages multi-turn credit assignment to identify key decision points in interaction trajectories, converting raw experiences into structured, retrievable knowledge, and design a memory evolution mechanism with mutation, generalization, crossover, and pruning operators that refines the memory library through meta-level optimization.
    \item We demonstrate that \textit{PRIME} achieves competitive performance with RL-trained agents on eight user-centric benchmarks while requiring much fewer GPU-hours, and that memory libraries transfer effectively across model architectures.
\end{itemize}

\section{Preliminaries}

\paragraph{Multi-Turn User-Agent Interaction}

We model multi-turn agent-user interaction~\cite{wu2025collabllm, yao2024tau, barres2025tau, qian2025userrl} as a Markov Decision Process $\mathcal{M} = (\mathcal{S}, \mathcal{A}, \mathcal{T}, \mathcal{R}, \gamma, H)$ where states $\mathcal{S}$ encompass conversation history and environment context, actions $\mathcal{A}$ combine structured response types with natural language, and horizon $H$ bounds interaction length. The agent executes policy $\pi_\theta(a|s)$ to produce trajectories $\tau = \{(s_t, a_t, r_t)\}_{t=0}^{T}$, with the objective of maximizing expected cumulative reward:
\begin{equation}
\pi_\theta^* = \arg\max_{\pi_\theta} \mathbb{E}_{\tau \sim \pi_\theta} \left[ \sum_{t=0}^{T} \gamma^t r_t \right]
\label{eq:rl_objective}
\end{equation}

User-centric tasks present several major challenges: (i) user intent is latent and must be inferred through dialogue, (ii) feedback is often sparse with meaningful rewards arriving only at episode conclusion, (iii) and the state space grows unboundedly with conversation length.

\paragraph{Group Relative Policy Optimization}

GRPO~\cite{guo2025deepseek} addresses reward sparsity by computing advantages relative to a group of sampled responses. Given prompt $x$ and responses $\{y_1, \ldots, y_G\}$ with rewards $\{r_1, \ldots, r_G\}$, advantages are normalized within each group:
\begin{equation}
A_i = \frac{r_i - \text{mean}(\{r_j\})}{\text{std}(\{r_j\})}
\label{eq:grpo_advantage}
\end{equation}

The policy updates to favor high-advantage responses subject to a KL constraint:
\begin{equation}
\mathcal{L}_{\text{GRPO}}(\theta) = \mathbb{E}_{x, \{y_i\}} \left[ \sum_{i=1}^{G} A_i \log \pi_\theta(y_i | x) - \beta \, \text{KL}(\pi_\theta \| \pi_{\text{ref}}) \right]
\label{eq:grpo_loss}
\end{equation}

While GRPO and related RL methods have demonstrated strong performance on multi-turn user-centric tasks~\citep{qian2025userrl}, they require substantial computational resources---typically multi-GPU clusters running for days---and produce model weights that are architecture-specific and non-transferable. These constraints motivate our exploration of gradient-free alternatives that can achieve competitive performance while offering interpretability and cross-model portability.



\begin{figure*}[t]
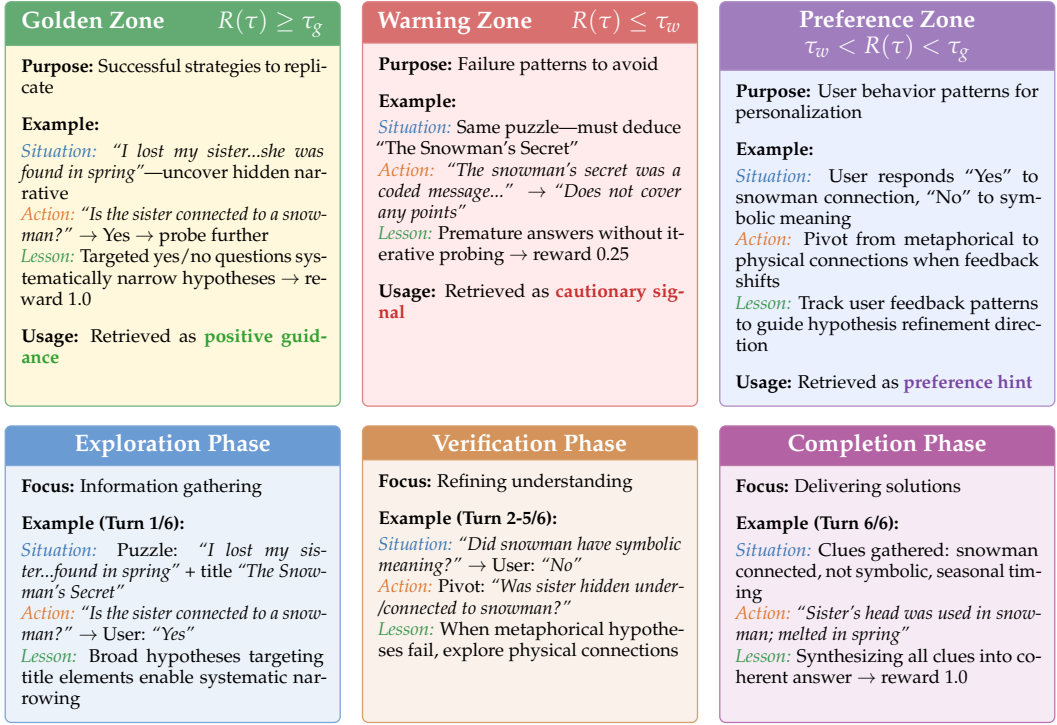

\centering

\begin{minipage}[t]{0.32\textwidth}
\begin{tcolorbox}[
    colback=goldenbg,
    colframe=headergreen!70,
    title={\centering\small\color{white} Golden Zone \hfill $R(\tau) \geq \tau_g$},
    fonttitle=\bfseries,
    colbacktitle=headergreen!80,
    coltitle=white,
    boxrule=0.5pt,
    left=3pt, right=3pt, top=2pt, bottom=2pt,
    arc=2pt,
    equal height group=fig4zones
]
\scriptsize
\textbf{Purpose:} Successful strategies to replicate

\medskip
\textbf{Example:}\\[2pt]
\textcolor{headerblue}{\textit{Situation:}} \textit{``I lost my sister...she was found in spring''}---uncover hidden narrative\\
\textcolor{headerorange}{\textit{Action:}} \textit{``Is the sister connected to a snowman?''} $\to$ Yes $\to$ probe further\\
\textcolor{headergreen}{\textit{Lesson:}} Targeted yes/no questions systematically narrow hypotheses $\to$ reward 1.0

\medskip
\textbf{Usage:} Retrieved as \textcolor{goodgreen}{\textbf{positive guidance}}
\end{tcolorbox}
\end{minipage}%
\hfill
\begin{minipage}[t]{0.32\textwidth}
\begin{tcolorbox}[
    colback=warningbg,
    colframe=badred!60,
    title={\centering\small\color{white} Warning Zone \hfill $R(\tau) \leq \tau_w$},
    fonttitle=\bfseries,
    colbacktitle=badred!70,
    coltitle=white,
    boxrule=0.5pt,
    left=3pt, right=3pt, top=2pt, bottom=2pt,
    arc=2pt,
    equal height group=fig4zones
]
\scriptsize
\textbf{Purpose:} Failure patterns to avoid

\medskip
\textbf{Example:}\\[2pt]
\textcolor{headerblue}{\textit{Situation:}} Same puzzle---must deduce ``The Snowman's Secret''\\
\textcolor{headerorange}{\textit{Action:}} \textit{``The snowman's secret was a coded message...''} $\to$ \textit{``Does not cover any points''}\\
\textcolor{headergreen}{\textit{Lesson:}} Premature answers without iterative probing $\to$ reward 0.25

\medskip
\textbf{Usage:} Retrieved as \textcolor{badred}{\textbf{cautionary signal}}
\end{tcolorbox}
\end{minipage}%
\hfill
\begin{minipage}[t]{0.32\textwidth}
\begin{tcolorbox}[
    colback=prefbg,
    colframe=headerpurple!60,
    title={\centering\small\color{white} Preference Zone \hfill $\tau_w < R(\tau) < \tau_g$},
    fonttitle=\bfseries,
    colbacktitle=headerpurple!70,
    coltitle=white,
    boxrule=0.5pt,
    left=3pt, right=3pt, top=2pt, bottom=2pt,
    arc=2pt,
    equal height group=fig4zones
]
\scriptsize
\textbf{Purpose:} User behavior patterns for personalization

\medskip
\textbf{Example:}\\[2pt]
\textcolor{headerblue}{\textit{Situation:}} User responds ``Yes'' to snowman connection, ``No'' to symbolic meaning\\
\textcolor{headerorange}{\textit{Action:}} Pivot from metaphorical to physical connections when feedback shifts\\
\textcolor{headergreen}{\textit{Lesson:}} Track user feedback patterns to guide hypothesis refinement direction

\medskip
\textbf{Usage:} Retrieved as \textcolor{headerpurple}{\textbf{preference hint}}
\end{tcolorbox}
\end{minipage}

\vspace{6pt}

\begin{minipage}[t]{0.32\textwidth}
\begin{tcolorbox}[
    colback=exploreclr!10,
    colframe=exploreclr!70,
    title={\centering\small\color{white} Exploration Phase},
    fonttitle=\bfseries,
    colbacktitle=exploreclr!80,
    coltitle=white,
    boxrule=0.5pt,
    left=3pt, right=3pt, top=2pt, bottom=2pt,
    arc=2pt,
    equal height group=fig4stages
]
\scriptsize
\textbf{Focus:} Information gathering

\medskip
\textbf{Example (Turn 1/6):}\\[2pt]
\textcolor{headerblue}{\textit{Situation:}} Puzzle: \textit{``I lost my sister...found in spring''} + title \textit{``The Snowman's Secret''}\\
\textcolor{headerorange}{\textit{Action:}} \textit{``Is the sister connected to a snowman?''} $\to$ User: \textit{``Yes''}\\
\textcolor{headergreen}{\textit{Lesson:}} Broad hypotheses targeting title elements enable systematic narrowing
\end{tcolorbox}
\end{minipage}%
\hfill
\begin{minipage}[t]{0.32\textwidth}
\begin{tcolorbox}[
    colback=distillclr!10,
    colframe=distillclr!70,
    title={\centering\small\color{white} Verification Phase},
    fonttitle=\bfseries,
    colbacktitle=distillclr!80,
    coltitle=white,
    boxrule=0.5pt,
    left=3pt, right=3pt, top=2pt, bottom=2pt,
    arc=2pt,
    equal height group=fig4stages
]
\scriptsize
\textbf{Focus:} Refining understanding

\medskip
\textbf{Example (Turn 2-5/6):}\\[2pt]
\textcolor{headerblue}{\textit{Situation:}} \textit{``Did snowman have symbolic meaning?''} $\to$ User: \textit{``No''}\\
\textcolor{headerorange}{\textit{Action:}} Pivot: \textit{``Was sister hidden under/connected to snowman?''}\\
\textcolor{headergreen}{\textit{Lesson:}} When metaphorical hypotheses fail, explore physical connections
\end{tcolorbox}
\end{minipage}%
\hfill
\begin{minipage}[t]{0.32\textwidth}
\begin{tcolorbox}[
    colback=inferclr!10,
    colframe=inferclr!70,
    title={\centering\small\color{white} Completion Phase},
    fonttitle=\bfseries,
    colbacktitle=inferclr!80,
    coltitle=white,
    boxrule=0.5pt,
    left=3pt, right=3pt, top=2pt, bottom=2pt,
    arc=2pt,
    equal height group=fig4stages
]
\scriptsize
\textbf{Focus:} Delivering solutions

\medskip
\textbf{Example (Turn 6/6):}\\[2pt]
\textcolor{headerblue}{\textit{Situation:}} Clues gathered: snowman connected, not symbolic, seasonal timing\\
\textcolor{headerorange}{\textit{Action:}} \textit{``Sister's head was used in snowman; melted in spring''}\\
\textcolor{headergreen}{\textit{Lesson:}} Synthesizing all clues into coherent answer $\to$ reward 1.0
\end{tcolorbox}
\end{minipage}

\caption{\textbf{Memory library organization.} \textit{Top row:} Memories are organized by trajectory reward $R(\tau)$ into three semantic zones---golden (successful strategies), warning (failure patterns), and preference (user behavior patterns). \textit{Bottom row:} Each memory is also annotated with its interaction stage---exploration (information gathering), verification (refining understanding), or completion (delivering solutions). Core components are color-coded: \textcolor{headerblue}{situation}, \textcolor{headerorange}{action}, and \textcolor{headergreen}{lesson}.}
\label{fig:zones}
\end{figure*}



\begin{figure*}[t]
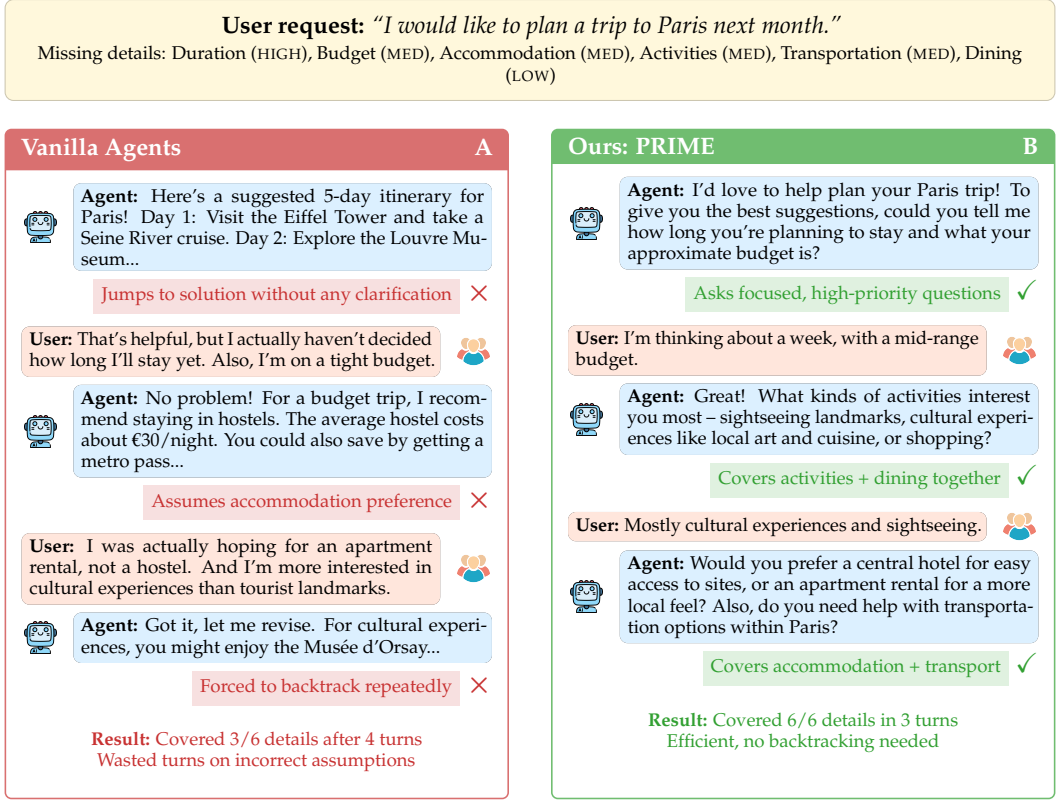

\centering

\begin{tcolorbox}[
    colback=goldenbg,
    colframe=goldenbg!80!black,
    boxrule=0.4pt,
    left=4pt, right=4pt, top=3pt, bottom=3pt,
    arc=3pt,
    width=\textwidth
]
\small\centering
\textbf{User request:} \textit{``I would like to plan a trip to Paris next month.''}\\[2pt]
\scriptsize Missing details: Duration (\textsc{high}), Budget (\textsc{med}), Accommodation (\textsc{med}), Activities (\textsc{med}), Transportation (\textsc{med}), Dining (\textsc{low})
\end{tcolorbox}

\vspace{4pt}

\begin{minipage}[t]{0.48\textwidth}
\begin{tcolorbox}[
    colback=white,
    colframe=badred!70,
    title={\centering\small Vanilla Agents \hfill \textbf{A}},
    fonttitle=\bfseries,
    boxrule=0.6pt,
    left=3pt, right=3pt, top=2pt, bottom=2pt,
    arc=2pt,
    equal height group=fig2compare
]

\agenticon{Here's a suggested 5-day itinerary for Paris! Day 1: Visit the Eiffel Tower and take a Seine River cruise. Day 2: Explore the Louvre Museum...}

\vspace{2pt}
\hfill\colorbox{badred!15}{\scriptsize\color{badred} Jumps to solution without any clarification} \badmark

\vspace{4pt}

\usericon{That's helpful, but I actually haven't decided how long I'll stay yet. Also, I'm on a tight budget.}

\vspace{2pt}

\agenticon{No problem! For a budget trip, I recommend staying in hostels. The average hostel costs about \texteuro 30/night. You could also save by getting a metro pass...}

\vspace{2pt}
\hfill\colorbox{badred!15}{\scriptsize\color{badred} Assumes accommodation preference} \badmark

\vspace{4pt}

\usericon{I was actually hoping for an apartment rental, not a hostel. And I'm more interested in cultural experiences than tourist landmarks.}

\vspace{2pt}

\agenticon{Got it, let me revise. For cultural experiences, you might enjoy the Mus\'ee d'Orsay...}

\vspace{2pt}
\hfill\colorbox{badred!15}{\scriptsize\color{badred} Forced to backtrack repeatedly} \badmark

\vspace{4pt}
\begin{center}
\scriptsize\color{badred} \textbf{Result:} Covered 3/6 details after 4 turns\\
Wasted turns on incorrect assumptions
\end{center}

\end{tcolorbox}
\end{minipage}%
\hfill
\begin{minipage}[t]{0.48\textwidth}
\begin{tcolorbox}[
    colback=white,
    colframe=goodgreen!70,
    title={\centering\small Ours: PRIME \hfill \textbf{B}},
    fonttitle=\bfseries,
    boxrule=0.6pt,
    left=3pt, right=3pt, top=2pt, bottom=2pt,
    arc=2pt,
    equal height group=fig2compare
]

\agenticon{I'd love to help plan your Paris trip! To give you the best suggestions, could you tell me how long you're planning to stay and what your approximate budget is?}

\vspace{2pt}
\hfill\colorbox{goodgreen!15}{\scriptsize\color{goodgreen} Asks focused, high-priority questions} \goodmark

\vspace{4pt}

\usericon{I'm thinking about a week, with a mid-range budget.}

\vspace{2pt}

\agenticon{Great! What kinds of activities interest you most -- sightseeing landmarks, cultural experiences like local art and cuisine, or shopping?}

\vspace{2pt}
\hfill\colorbox{goodgreen!15}{\scriptsize\color{goodgreen} Covers activities + dining together} \goodmark

\vspace{4pt}

\usericon{Mostly cultural experiences and sightseeing.}

\vspace{2pt}

\agenticon{Would you prefer a central hotel for easy access to sites, or an apartment rental for a more local feel? Also, do you need help with transportation options within Paris?}

\vspace{2pt}
\hfill\colorbox{goodgreen!15}{\scriptsize\color{goodgreen} Covers accommodation + transport} \goodmark

\vspace{4pt}
\begin{center}
\scriptsize\color{goodgreen} \textbf{Result:} Covered 6/6 details in 3 turns\\
Efficient, no backtracking needed
\end{center}

\end{tcolorbox}
\end{minipage}

\caption{\textbf{Vanilla LLM vs.\ PRIME on an IntentionGym task.} \textbf{(A)}~Without experience guidance, the agent jumps directly to providing solutions, makes incorrect assumptions about user preferences, and is forced to backtrack when corrected. \textbf{(B)}~With PRIME, the agent systematically asks focused clarifying questions that efficiently cover all missing details, guided by retrieved experiences from similar past interactions. Green checks (\goodmark) mark productive turns that address user needs; red crosses (\badmark) mark turns that assume rather than ask.}
\label{fig:comparison}
\end{figure*}


\section{Method}

We present \textit{PRIME (\textbf{P}roactive \textbf{R}easoning via \textbf{I}terative \textbf{M}emory \textbf{E}volution)} (Figure~\ref{fig:system_overview}), a gradient-free learning framework that enables LLM agents to continuously improve their user-centric interaction abilities through accumulated interaction memories, without modifying model parameters. Unlike gradient-based reinforcement learning methods that require expensive multi-GPU training, PRIME constructs an interpretable, transferable experience library that captures successful interaction strategies and failure patterns from multi-turn dialogues.

\subsection{Learning as Memory Evolution}

We formalize PRIME as an optimization problem over memory libraries rather than model parameters. Given a data distribution $\mathcal{D}$ over input-target pairs $(X, Y)$, a frozen LLM policy $\pi$, and a retrieval function $\rho$ that selects relevant memories from the library, the objective is to find an optimal memory library $\mathcal{M}^*$ that maximizes expected agent performance:
\begin{equation}
\mathcal{M}^*= \arg\max_{\mathcal{M}} \mathbb{E}_{(X_i, Y_i) \sim \mathcal{D}, \bold{m_i} \sim \rho(\cdot|X_i, \mathcal{M})} \left[ \Phi(\pi(\cdot | X_i, \bold{m_i}), Y_i) \right]
\label{eq:objective}
\end{equation}

Here, the performance metric $\Phi$ measures how well the policy's output matches the desired target. $X_i$ represents the input context comprising user queries, environment state, and conversation history, while $Y_i$ captures the target outcome including task completion and user satisfaction. The retrieval function $\rho(\cdot|X_i, \mathcal{M})$ selects experiences $m_i$ from library $\mathcal{M}$ that are semantically relevant to the current situation. The policy $\pi(\cdot|X_i, \bold{m_i})$ is the frozen LLM conditioned on both the input and retrieved experiences.

The key distinction from gradient-based methods~\cite{qian2025userrl, wu2025collabllm} is that we optimize the memory library $\mathcal{M}$ rather than the policy parameters $\theta$. This optimization proceeds through a forward update mechanism where the library evolves based on collected interaction trajectories:
\begin{equation}
\mathcal{M}_{i+1} \sim \mu(\cdot | \mathcal{M}_i, \{\tau_i | X_i, \pi\})
\label{eq:evolution}
\end{equation}

The updater distribution $\mu$ incorporates new experiences distilled from trajectories $\tau_i$ and applies evolution operations to refine existing knowledge. We can interpret this as a \textit{forward gradient} that represents the direction of library improvement without requiring backpropagation through the policy.

\paragraph{Structuring Interaction Knowledge}

In the memory library, each memory $\bold{m} \in \mathcal{M}$ captures interaction knowledge in a structured representation with rich contextual metadata for stage-aware retrieval:
\begin{equation}
m = (m_{\text{core}}, m_{\text{stage}}, m_{\text{cond}}, m_{\text{zone}})
\end{equation}
where $m_{\text{core}} = (\textcolor{headerblue}{\textit{situation}}, \textcolor{headerorange}{\textit{action}}, \textit{outcome}, \textcolor{headergreen}{\textit{lesson}})$ captures the interaction knowledge, $m_{\text{stage}}$ records the interaction phase, $m_{\text{cond}}$ specifies applicability conditions, and $m_{\text{zone}}$ indicates the semantic zone classification. Figure~\ref{fig:zones} illustrates each component with concrete examples.

The \textit{core} components capture what happened during the interaction. For instance, in a travel planning scenario, the \textcolor{headerblue}{\textit{situation}} might be ``user has vague multi-aspect planning request,'' the \textcolor{headerorange}{\textit{action}} could be ``pair related questions (venue + headcount, budget + theme),'' the \textit{outcome} records the user's positive response, and the \textcolor{headergreen}{\textit{lesson}} abstracts the principle ``combining related details in single questions covers ground efficiently.''

The \textit{stage} field records the interaction phase in which the memory occurred. We define three canonical phases: the \textbf{exploration phase} (early interaction focused on information gathering), the \textbf{verification phase} (middle portion for refining understanding), and the \textbf{completion phase} (final stage for delivering solutions). The bottom row of Figure~\ref{fig:zones} illustrates each phase with examples. The \textit{conditions} field specifies applicability criteria such as environment types and preconditions for retrieval.

We organize memories into three semantic zones based on trajectory rewards (Figure~\ref{fig:zones}): the \textit{Golden Zone} contains successful strategies to replicate, the \textit{Warning Zone} captures failure patterns to avoid (e.g., ``jumped to generating full itinerary without clarifying questions'' leading to repeated backtracking), and the \textit{Preference Zone} records user behavior patterns that enable personalization. This organization ensures positive guidance dominates during retrieval while cautionary signals remain available.

\subsection{The PRIME Pipeline: Exploration to Inference}

The PRIME pipeline enables progressive improvement of the memory library and consequently of agent performance through three phases: (i) \textit{exploration}, (ii) \textit{distillation}, and (iii) \textit{evolution}.

\vspace{-6pt}
\paragraph{Exploration: Collecting Interaction Trajectories}

The exploration phase generates raw interaction data. For each episode, the agent interacts with users multi-turn conversations, producing trajectories of the form $\tau = \{(s_t, a_t, r_t, o_t)\}_{t=0}^{T}$ where $s_t$ represents the state at turn $t$ comprising conversation history, environment context, and any retrieved memories; $a_t$ is the agent's action combining structured action types with natural language content; $r_t$ captures turn-level rewards; and $o_t$ records environment observation.

When memory guidance is enabled, the policy incorporates retrieved knowledge through prompt augmentation:
\begin{equation}
\pi(a_t | s_t, \mathcal{M}) = \pi_{\text{LLM}}(a_t | \text{prompt}(s_t, \rho(s_t, \mathcal{M})))
\end{equation}

During exploration, we collect trajectories that cover both successful and unsuccessful interaction patterns. This diversity is essential for building a comprehensive memory library that captures the full range of situations the agent might encounter.

\paragraph{Distillation: From Trajectories to Memories}

The distillation phase transforms raw trajectories into structured experiences with rich contextual metadata for stage-aware retrieval. This process must identify which aspects of an interaction are worth preserving, determine the interaction stage where critical decisions occurred, and compress the trajectory into a representation that supports contextualized retrieval at inference time.

The first step is \textit{credit assignment}, which identifies which turns contributed most to the final outcome. Given turn-level credits, we select the top-$K$ turns as \textit{key turns} that capture the most informative moments of the interaction. The interaction stage $\sigma$ is then determined by where these key turns occurred in the trajectory. Figure~\ref{fig:distillation} illustrates this process with a concrete example.

The distillation function then leverages an LLM to extract semantic knowledge:
\begin{equation}
m = \mathcal{D}(\tau, \mathcal{K}, \sigma) = \text{LLM}_{\text{distill}}(\tau, \mathcal{K}, \sigma)
\end{equation}
where $\tau$ is the current interaction trajectory, $\mathcal{K}$ denotes the key turns, and $\sigma$ is the detected stage. For example, from a successful travel planning trajectory where the agent efficiently gathered user preferences through paired questions (Figure~\ref{fig:distillation}), the distiller extracts: $m_{\text{core}}$ capturing the strategy of combining related questions, $m_{\text{stage}} = \text{exploration}$ since key turns occurred early, $m_{\text{cond}}$ specifying applicability to multi-aspect planning tasks, and $m_{\text{zone}} = \text{golden}$ based on the high trajectory reward.

\paragraph{Evolution: Refining the Knowledge Base}

The memory library evolves through meta-level optimization, instantiating the update distribution $\mu$ from Equation~\ref{eq:evolution}. Four operators modify the library $\mathcal{M}_{i} \to \mathcal{M}_{i+1}$ (Figure~\ref{fig:evolution}):
\begin{itemize}[leftmargin=*, nosep, topsep=2pt]
    \item \textbf{Mutation}: Refines experience clarity based on empirical feedback.
    \item \textbf{Generalization}: Abstracts environment-specific details from high-performing experiences.
    \item \textbf{Crossover}: Combines complementary insights from multiple experiences.
    \item \textbf{Pruning}: Removes stale or underperforming experiences from the library.
\end{itemize}

\vspace{-10pt}

\paragraph{Memory-Guided Inference}

At inference time, the evolved library guides agent responses through \textit{contextualized retrieval}. PRIME formats candidate experiences as callable tools and asks the agent to select the most relevant ones given the current interaction state and stage:
\begin{equation}
\mathbf{m} = \rho(s, \mathcal{M}, C)
\end{equation}
where $s$ is the current state, $\mathcal{M}$ is the memory library, and $C = (h, t, H)$ captures the conversation history $h$, current turn $t$, and maximum turns $H$. Selected experiences are organized by zone in the augmented prompt: golden experiences provide positive guidance, warning experiences highlight pitfalls to avoid, and preference experiences capture user behavior patterns. Due to space constraints, we defer the details of the retrieval pipeline to Appendix (Figure~\ref{fig:retrieval}).

\section{Experiments}

We evaluate PRIME across diverse multi-turn, user-centric agentic tasks, comparing against both prompting-based and RL-trained baselines.

\begin{figure}[t]
    \centering
    \includegraphics[width=\linewidth]{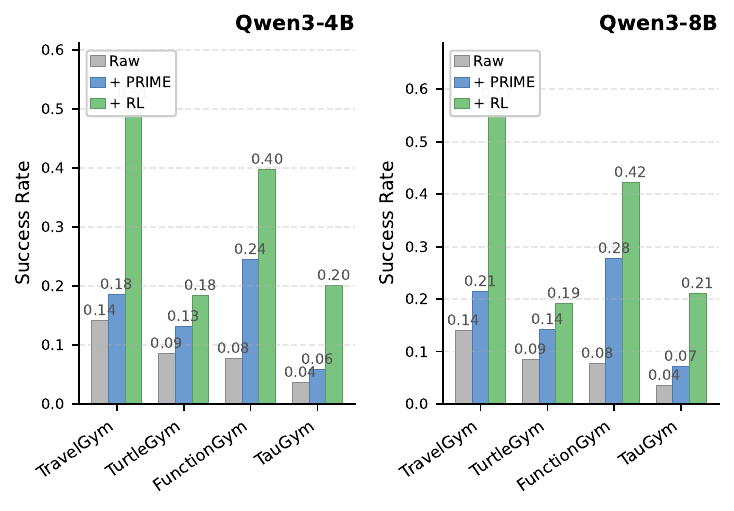}
    \caption{Comparison between base Raw models, PRIME, and RL-based models across challenging user-centric environments.}
    \label{fig:main_results}
\end{figure}

\subsection{Experimental Setup}

\paragraph{Benchmarks}
We evaluate on eight multi-turn, user-centric environments adopted from the UserRL benchmark~\cite{qian2025userrl}, each posing a distinct interaction challenge. \textbf{TurtleGym} requires lateral thinking: the agent must solve ``Turtle Soup'' puzzles by asking strategic yes/no questions to uncover a hidden narrative. \textbf{TelepathyGym} tests logical reasoning through an entity-guessing game where the agent narrows down possibilities via elimination. \textbf{PersuadeGym} measures persuasive communication, tasking the agent with shifting an interlocutor's stance through argumentation. \textbf{IntentionGym} evaluates intent inference, where the agent must discover underspecified user goals through clarifying questions. \textbf{FunctionGym} targets mathematical reasoning by requiring the agent to identify a hidden function from input--output examples. \textbf{SearchGym} tests information retrieval, combining web search with multi-step synthesis to answer factual queries. \textbf{TauGym}~\cite{yao2024tau} presents tool-use scenarios modeled after customer support, requiring multi-agent coordination with external APIs. Finally, \textbf{TravelGym} involves preference elicitation for trip planning, where the agent must proactively uncover latent user preferences. All environments share a standardized action interface and use LLM-based user simulation, enabling consistent evaluation across domains.

\vspace{-6pt}
\paragraph{Baselines}

We compare PRIME against three categories of baselines. First, \textit{proprietary LLMs}: we evaluate GPT-4o~\cite{achiam2023gpt} under zero-shot prompting to establish the performance ceiling of large-scale models without task-specific adaptation. Second, \textit{prompting-based agents}: we test ReAct~\cite{yao2022react}, which interleaves reasoning traces with actions, and Reflexion~\cite{shinn2023reflexion}, which augments agents with verbal self-reflection on past failures. These baselines represent established strategies for improving agent behavior without parameter updates and serve as the most direct comparison to PRIME's experience-guided approach. Third, \textit{RL-trained agents}: we include models fine-tuned with GRPO using multi-turn credit assignment~\cite{qian2025userrl}.

\vspace{-12pt}
\paragraph{Training Details}

We use Qwen3-4B and Qwen3-8B~\cite{yang2025qwen3} as our primary base models for both PRIME and the RL baselines. For PRIME, the experience library is initialized empty and built through iterative exploration--distillation--evolution cycles. Exploration uses a rollout temperature of $\tau = 1.0$ with a maximum horizon of $H = 16$ turns per episode, matching the GRPO training configuration. The discount factor is set to $\gamma = 0.8$ for both reward-to-go credit assignment and trajectory reward computation. Zone classification thresholds are $\theta_{\text{golden}} = 0.7$ and $\theta_{\text{warning}} = 0.3$. For distillation and evolution, we use GPT-4o as the summarization and reasoning backbone. Contextualized retrieval presents up to 20 candidate experiences as tools to the selection LLM, which returns the most relevant subset given the current conversation state and interaction stage. For the RL baselines, we train with GRPO following the UserRL configuration: 8$\times$A100 GPUs, group size $G = 8$, and KL coefficient $\beta = 0.001$. All environments use Qwen3-32B as the user simulator to ensure consistent and reproducible evaluation across methods. The full set of hyperparameters is provided in Appendix~\ref{sec:appendix_hyperparams}.

\subsection{Experimental Results}

\paragraph{Evaluation on User-centric Multi-turn Benchmarks} Table~\ref{tab:main_results} presents results across eight user-centric multi-turn agent benchmarks, comparing closed-source models, open-source models, GRPO-trained models, and PRIME-augmented models with the learned memory library. PRIME consistently improves over raw open-source models across all environments and model scales, with the largest gains on environments requiring strategic exploration such as FunctionGym, TauGym, and TurtleGym. The memory library effectively captures transferable interaction strategies that benefit the frozen base model without any parameter updates. Compared to RL-based models such as those trained via GRPO, PRIME shows competitive performance while requiring no expensive gradient-based training. GRPO training produces larger absolute performance gains through direct parameter optimization, while PRIME operates without gradient updates or multi-GPU training---making it applicable when model weights cannot be modified. Figure~\ref{fig:main_results} visualizes performance on challenging environments where PRIME significantly outperforms base models; for instance, Qwen3-8B + PRIME achieves 193\% improvement on FunctionGym and 50\% on TauGym over the raw base model.

\begin{table*}[t]
\centering
\caption{\textbf{Main results on multi-turn user-interacting benchmarks.} We report task completion scores across eight user-centric environments (higher is better). Gray rows indicate PRIME (ours). {\color{blue}Blue values in parentheses} show the percentage improvement over the corresponding raw open-source model.}
\label{tab:main_results}
\resizebox{\textwidth}{!}{%
\begin{tabular}{@{}l|cccccccc@{}}
\toprule
\textbf{Model} & \textbf{TravelGym} & \textbf{TurtleGym} & \textbf{FunctionGym} & \textbf{TauGym} & \textbf{PersuadeGym} & \textbf{IntentionGym} & \textbf{TelepathyGym} & \textbf{SearchGym} \\
\midrule
\multicolumn{9}{c}{\textit{Closed-Source LLM}} \\
\midrule
GPT-4o & 0.364 & 0.292 & 0.282 & 0.030 & 0.377 & 1.898 & 0.854 & 0.880  \\
GPT-4o-mini & 0.098 & 0.091 & 0.154 & 0.206 & 0.532 & 0.250 & 0.049 & 0.352  \\
Gemini-2.5-Pro & 0.347 & 0.274 & 0.410 & 0.194 & 0.425 & 1.590 & 0.902 & 0.928  \\
Gemini-2.5-Flash & 0.255 & 0.196 & 0.321 & 0.121 & 0.409 & 1.685 & 0.634 & 0.928  \\
\midrule
\multicolumn{9}{c}{\textit{Open-Source LLM}} \\
\midrule
Qwen3-32B & 0.172 & 0.151 & 0.154 & 0.000 & 0.484 & 1.830 & 0.561 & 0.792  \\
Qwen3-14B & 0.192 & 0.142 & 0.167 & 0.103 & 0.532 & 1.700 & 0.585 & 0.512  \\
Qwen3-8B & 0.158 & 0.098 & 0.095 & 0.048 & 0.441 & 1.725 & 0.510 & 0.808  \\
Qwen3-4B & 0.141 & 0.085 & 0.077 & 0.036 & 0.405 & 1.740 & 0.488 & 0.856  \\
\midrule
\multicolumn{9}{c}{\textit{RL (GRPO) Trained Models}} \\
\midrule
GRPO $+$ Qwen3-4B & 0.509 & 0.184 & 0.397 & 0.200 & 0.579 & 1.808 & 0.634 & 0.864 \\
GRPO $+$ Qwen3-8B & 0.573 & 0.192 & 0.423 & 0.210 & 0.532 & 1.903 & 0.561 & 0.888 \\
\midrule
\multicolumn{9}{c}{\textit{PRIME Models}} \\
\midrule
\rowcolor{gray!20}
Qwen3-4B + PRIME & 0.185 {\footnotesize\color{blue}($\uparrow$31\%)} & 0.132 {\footnotesize\color{blue}($\uparrow$55\%)} & 0.245 {\footnotesize\color{blue}($\uparrow$218\%)} & 0.058 {\footnotesize\color{blue}($\uparrow$61\%)} & 0.498 {\footnotesize\color{blue}($\uparrow$23\%)} & 1.772 {\footnotesize\color{blue}($\uparrow$1.8\%)} & 0.558 {\footnotesize\color{blue}($\uparrow$14\%)} & 0.860 {\footnotesize\color{blue}($\uparrow$0.5\%)} \\
\rowcolor{gray!20}
Qwen3-8B + PRIME & 0.215 {\footnotesize\color{blue}($\uparrow$36\%)} & 0.142 {\footnotesize\color{blue}($\uparrow$45\%)} & 0.278 {\footnotesize\color{blue}($\uparrow$193\%)} & 0.072 {\footnotesize\color{blue}($\uparrow$50\%)} & 0.472 {\footnotesize\color{blue}($\uparrow$7.0\%)} & 1.825 {\footnotesize\color{blue}($\uparrow$5.8\%)} & 0.524 {\footnotesize\color{blue}($\uparrow$2.7\%)} & 0.872 {\footnotesize\color{blue}($\uparrow$7.9\%)} \\
\rowcolor{gray!20}
Qwen3-14B + PRIME & 0.248 {\footnotesize\color{blue}($\uparrow$29\%)} & 0.168 {\footnotesize\color{blue}($\uparrow$18\%)} & 0.298 {\footnotesize\color{blue}($\uparrow$78\%)} & 0.124 {\footnotesize\color{blue}($\uparrow$20\%)} & 0.568 {\footnotesize\color{blue}($\uparrow$6.8\%)} & 1.782 {\footnotesize\color{blue}($\uparrow$4.8\%)} & 0.632 {\footnotesize\color{blue}($\uparrow$8.0\%)} & 0.892 {\footnotesize\color{blue}($\uparrow$35\%)} \\
\bottomrule
\end{tabular}%
}
\end{table*}

\begin{figure}[t]
    \centering
    \begin{subfigure}[t]{0.48\textwidth}
        \centering
        \includegraphics[width=\linewidth]{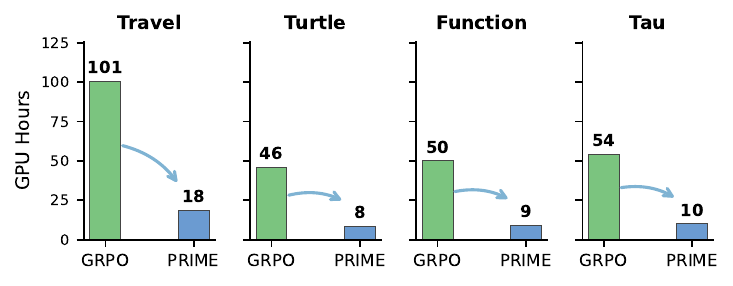}
        \caption{GRPO requires gradient-based training; PRIME uses inference-time exploration, achieving 5--6$\times$ reduction in GPU-hours.}
        \label{fig:compute}
    \end{subfigure}%
    \hfill
    \begin{subfigure}[t]{0.48\textwidth}
        \centering
        \includegraphics[width=\linewidth]{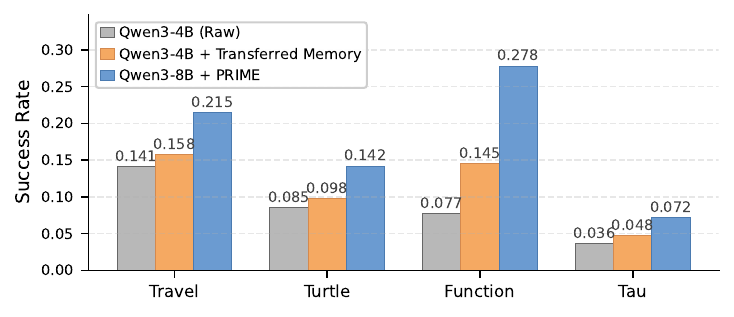}
        \caption{Qwen3-4B with a transferred 8B library (orange) improves over the raw 4B baseline (gray).}
        \label{fig:transfer}
    \end{subfigure}
    \caption{\textbf{Efficiency and transferability.}}
    \label{fig:compute_and_transfer}
\end{figure}

\paragraph{Computational Budget} Figure~\ref{fig:compute} compares the per-environment computational requirements of GRPO and PRIME. GRPO requires gradient-based optimization with multi-GPU training: 8$\times$A100 GPUs for approximately over 100 GPU hours. In contrast, PRIME eliminates gradient computation entirely---the exploration, distillation and evolution phases require only forward passes through the frozen base model. This architectural difference yields substantial savings: PRIME requires 5--6 times fewer GPU-hours per environment compared to GRPO. The trade-off is that GRPO achieves higher absolute performance through direct parameter updates, while PRIME provides a gradient-free path to improving agent behavior that is particularly attractive when training infrastructure is limited or model weights cannot be modified.

\vspace{-6pt}
\paragraph{Cross-Architecture Transferability}

A distinguishing property of experience-based learning is that the knowledge stored in the library is decoupled from model architecture and parameters. To evaluate this, we construct experience libraries using Qwen3-8B and apply them to Qwen3-4B at inference time. Figure~\ref{fig:transfer} shows the results on four challenging environments. Applying an 8B-derived library to 4B yields modest but consistent improvements over the raw baseline, though the gains are smaller than those achieved by native PRIME on the source 8B model. This gap is expected: experiences distilled from a larger model's successful trajectories may encode reasoning patterns that are less natural for the smaller model to execute. Nevertheless, the transferred library provides meaningful improvements without any gradient-based training on the 4B model. This has practical implications: organizations can build experience libraries using capable models during development, then deploy smaller models augmented with those libraries in production---establishing experience libraries as reusable assets that provide some benefit even when applied to different model architectures.

\section{Related Work}

\paragraph{Tool-Augmented Agents for User Interaction}

The deployment of LLM agents in user-facing applications has spurred development of benchmarks and frameworks for evaluating multi-turn, tool-using interactions. $\tau$-Bench~\cite{yao2024tau, barres2025tau} introduces realistic customer service scenarios requiring agents to coordinate multiple tools while maintaining coherent dialogue. UserBench~\cite{qian2025userbench} provides a comprehensive suite of user-centric environments spanning intent discovery, persuasion, and preference elicitation. These benchmarks reveal that current LLMs struggle with proactive information gathering---agents often fail to ask clarifying questions and instead make premature assumptions about user intent. CollabLLM~\cite{wu2025collabllm} addresses this by training agents to collaborate with users through iterative refinement. PRIME complements these efforts by providing a gradient-free mechanism for accumulating interaction strategies that can adapt to diverse user-centric scenarios.

\vspace{-6pt}
\paragraph{Reinforcement Learning for LLM Agents}

RL has emerged as a powerful paradigm for improving LLM capabilities beyond supervised learning. Early work applied RLHF~\cite{bai2022training} and DPO~\cite{rafailov2023direct} to align models with human preferences, while PPO~\cite{schulman2017proximal} enabled direct reward optimization. Recent advances include GRPO~\cite{guo2025deepseek}, which uses group-relative advantages to stabilize training, and variants like GiGPO~\cite{feng2025group} and IGPO~\cite{wang2025information} that improve sample efficiency. For multi-turn agent tasks, Turn-PPO~\cite{li2025turn} and RLVMR~\cite{zhang2025rlvmr} address credit assignment across extended interactions. UserRL~\cite{qian2025userrl} specifically targets user-centric tool-calling by combining GRPO with multi-turn credit assignment methods. While these RL approaches achieve strong performance, they require expensive multi-GPU training infrastructure and produce opaque parameter updates. PRIME offers an orthogonal approach: rather than optimizing model parameters, we optimize an explicit memory library that augments frozen models, achieving competitive performance with significantly lower computational cost and full interpretability.

\vspace{-6pt}
\paragraph{Agent Memories}

Equipping agents with external memory has a rich history in AI, from early cognitive architectures to modern retrieval-augmented generation. Recent work has explored learnable memory modules for LLM agents: MemGPT~\cite{xu2025mem} introduces hierarchical memory management for extended conversations, while MemoryAgent~\cite{yu2025memagent} uses RL to train memory read/write policies. PRIME extends this line of work to user-centric tasks, building proactive agents that refine their memories through meta-level optimization without gradient computation. This design enables PRIME to accumulate transferable interaction strategies that improve agent performance without modifying model weights.

\section{Conclusion}

We introduced PRIME, a gradient-free framework for improving user-centric LLM agents through iterative memory evolution. By distilling multi-turn interaction trajectories into structured, human-readable experiences and evolving these memories through meta-level operators, PRIME enables frozen language models to progressively improve their interaction strategies without parameter updates. Our experiments across eight diverse user-centric benchmarks demonstrate that PRIME achieves competitive performance with RL-trained agents while requiring significantly fewer computational resources, and that memory libraries transfer across model architectures. PRIME opens a practical pathway toward deployable, interpretable agents that learn from experience---complementing gradient-based methods with a lightweight, auditable alternative.





\bibliography{example_paper}
\bibliographystyle{colm2026_conference}


\newpage
\appendix

\section{Memory Evolution Operators}
\label{sec:appendix_evolution}


\begin{figure*}[h]
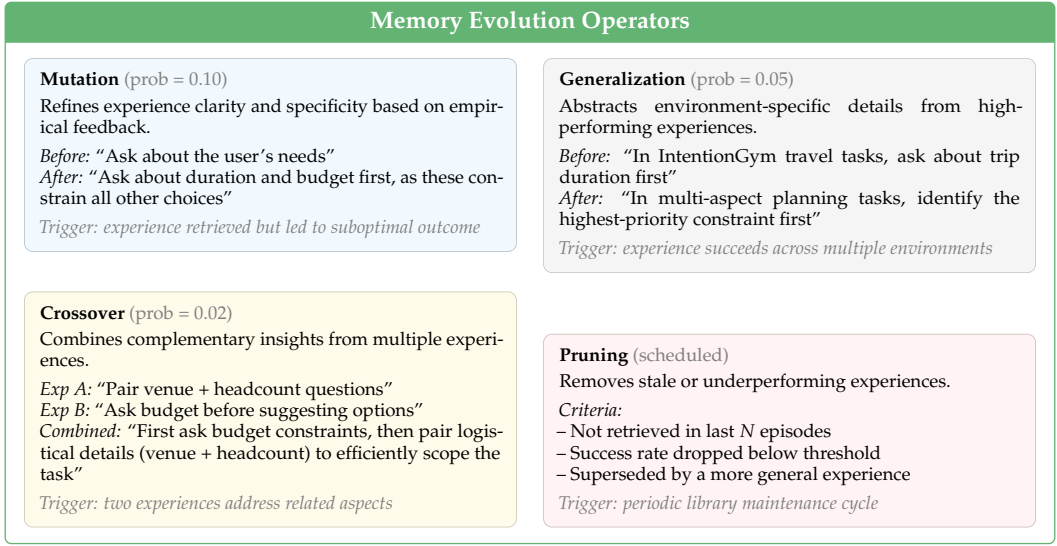

\centering
\begin{tcolorbox}[
    colback=white,
    colframe=evolveclr!80,
    title={\centering\small Memory Evolution Operators},
    fonttitle=\bfseries,
    colbacktitle=evolveclr!80,
    coltitle=white,
    boxrule=0.5pt,
    left=4pt, right=4pt, top=3pt, bottom=3pt,
    arc=2pt,
    width=\textwidth
]

\begin{multicols}{2}

\begin{tcolorbox}[colback=agentbg!40, boxrule=0.3pt, colframe=agentbg!80!black, arc=3pt, left=3pt, right=3pt, top=2pt, bottom=2pt]
\scriptsize
\textbf{Mutation} {\color{gray}(prob = 0.10)}\\[2pt]
Refines experience clarity and specificity based on empirical feedback.\\[3pt]
\textit{Before:} ``Ask about the user's needs''\\
\textit{After:} ``Ask about duration and budget first, as these constrain all other choices''\\[3pt]
{\color{gray}\textit{Trigger: experience retrieved but led to suboptimal outcome}}
\end{tcolorbox}

\vspace{4pt}

\begin{tcolorbox}[colback=goldenbg!60, boxrule=0.3pt, colframe=goldenbg!80!black, arc=3pt, left=3pt, right=3pt, top=2pt, bottom=2pt]
\scriptsize
\textbf{Crossover} {\color{gray}(prob = 0.02)}\\[2pt]
Combines complementary insights from multiple experiences.\\[3pt]
\textit{Exp A:} ``Pair venue + headcount questions''\\
\textit{Exp B:} ``Ask budget before suggesting options''\\
\textit{Combined:} ``First ask budget constraints, then pair logistical details (venue + headcount) to efficiently scope the task''\\[3pt]
{\color{gray}\textit{Trigger: two experiences address related aspects}}
\end{tcolorbox}

\columnbreak

\begin{tcolorbox}[colback=simbg, boxrule=0.3pt, colframe=simbg!80!black, arc=3pt, left=3pt, right=3pt, top=2pt, bottom=2pt]
\scriptsize
\textbf{Generalization} {\color{gray}(prob = 0.05)}\\[2pt]
Abstracts environment-specific details from high-performing experiences.\\[3pt]
\textit{Before:} ``In IntentionGym travel tasks, ask about trip duration first''\\
\textit{After:} ``In multi-aspect planning tasks, identify the highest-priority constraint first''\\[3pt]
{\color{gray}\textit{Trigger: experience succeeds across multiple environments}}
\end{tcolorbox}

\vspace{4pt}

\begin{tcolorbox}[colback=warningbg!60, boxrule=0.3pt, colframe=warningbg!80!black, arc=3pt, left=3pt, right=3pt, top=2pt, bottom=2pt]
\scriptsize
\textbf{Pruning} {\color{gray}(scheduled)}\\[2pt]
Removes stale or underperforming experiences.\\[3pt]
\textit{Criteria:}\\
-- Not retrieved in last $N$ episodes\\
-- Success rate dropped below threshold\\
-- Superseded by a more general experience\\[3pt]
{\color{gray}\textit{Trigger: periodic library maintenance cycle}}
\end{tcolorbox}

\end{multicols}

\end{tcolorbox}

\caption{\textbf{Memory evolution operators.} The library evolves through four meta-level operations applied probabilistically. \textit{Mutation} sharpens vague experiences based on feedback. \textit{Generalization} abstracts domain-specific successes into transferable knowledge. \textit{Crossover} synthesizes complementary insights into richer experiences. \textit{Pruning} removes stale entries to keep the library focused. Together, these operators enable the experience library to improve over time without gradient computation.}
\label{fig:evolution}
\end{figure*}

\section{Additional Distillation Examples}
\label{sec:appendix_distillation}

We provide additional examples of experience distillation from other UserRL environments beyond IntentionGym (shown in Figure~\ref{fig:distillation}).


\begin{figure*}[h]
\centering
\begin{tcolorbox}[
    colback=white,
    colframe=headerblue!80,
    title={\small Distillation Example \hspace{4pt}$|$\hspace{4pt} TurtleGym (Lateral Thinking)},
    fonttitle=\bfseries,
    colbacktitle=headerblue!80,
    coltitle=white,
    boxrule=0.5pt,
    left=4pt, right=4pt, top=3pt, bottom=3pt,
    arc=2pt,
    width=\textwidth
]

\small

\begin{tabular}{@{}p{2.2cm}p{\dimexpr\linewidth-2.2cm-2\tabcolsep\relax}@{}}
\rowcolor{agentbg!60}
\textbf{Task:} & Surface: ``I lost my sister, and it wasn't until the following spring that she was found...'' Goal: Explain what really happened in this story. \\[4pt]

\rowcolor{userbg!60}
\textbf{Trajectory:} &
\textbf{Turn 1:} [action] Is the sister connected to a snowman in any way? $\rightarrow$ ``Yes.'' ~~(r=0.6)\\
& \textbf{Turn 2:} [action] Did the sister become part of the snowman? $\rightarrow$ ``Yes.'' ~~(r=0.8) \\
& \textbf{Turn 3:} [action] Was her disappearance intentional? $\rightarrow$ ``Yes.'' ~~(r=0.4) \\
& \textbf{Turn 4:} [answer] The protagonist hid the sister inside the snowman... $\rightarrow$ \textit{(Correct)} ~~$R(\tau) = 1.00$ \\[4pt]

\rowcolor{goldenbg!80}
\textbf{Credit Assignment:} & R2G: $c = [0.15, 0.35, 0.20, 0.30]$. ~~ Key turns: \{1, 2\} (hypothesis-confirming questions) \\[4pt]

\rowcolor{agentbg!30}
\textbf{Zone:} & \textsc{Golden} ($R(\tau) = 1.00 \geq 0.7$) \\[4pt]

\rowcolor{simbg}
\textbf{Distilled Experience:} & \textbf{Situation:} Lateral thinking puzzle with a cryptic surface story containing symbolic elements \\
& \textbf{Action:} Start with broad categorical questions linking unusual elements (e.g., ``snowman''), then narrow via binary confirmation \\
& \textbf{Lesson:} Identify the most unusual/symbolic element in the surface story and form hypotheses around it; use yes/no questions to efficiently converge on the hidden narrative \\
& \textbf{Stage:} Exploration $\to$ Verification \\
& \textbf{Conditions:} \{env: [turtle], turn\_range: [1, 5], preconditions: [cryptic puzzle, symbolic elements]\} \\
\end{tabular}

\end{tcolorbox}
\caption{\textbf{Experience distillation from TurtleGym.} The agent identifies the key symbolic element (snowman) in turn 1 and uses binary questions to confirm the hypothesis. Credit assignment via R2G identifies turns 1-2 as pivotal. The distilled lesson---focusing on unusual elements and using binary confirmation---generalizes to other lateral thinking puzzles.}
\label{fig:distillation_turtle}
\end{figure*}


\begin{figure*}[h]
\centering
\begin{tcolorbox}[
    colback=white,
    colframe=headerblue!80,
    title={\small Distillation Example \hspace{4pt}$|$\hspace{4pt} TravelGym (Preference Elicitation)},
    fonttitle=\bfseries,
    colbacktitle=headerblue!80,
    coltitle=white,
    boxrule=0.5pt,
    left=4pt, right=4pt, top=3pt, bottom=3pt,
    arc=2pt,
    width=\textwidth
]

\small

\begin{tabular}{@{}p{2.2cm}p{\dimexpr\linewidth-2.2cm-2\tabcolsep\relax}@{}}
\rowcolor{agentbg!60}
\textbf{Task:} & User wants to plan a family vacation but hasn't specified destination, budget, dates, or activities. \\[4pt]

\rowcolor{userbg!60}
\textbf{Trajectory:} &
\textbf{Turn 1:} [action] What's your ideal climate---beach warmth, mountain cool, or city vibes? $\rightarrow$ ``Beach sounds nice, we haven't been to one in years.'' ~~(r=0.5)\\
& \textbf{Turn 2:} [action] How many days are you planning, and do kids need to be back for school? $\rightarrow$ ``About 5 days, mid-July when school's out.'' ~~(r=0.6) \\
& \textbf{Turn 3:} [action] What's your budget range, and do you prefer all-inclusive or flexible dining? $\rightarrow$ ``Around \$4000 total, flexible is better---the kids are picky.'' ~~(r=0.7) \\
& \textbf{Turn 4:} [search] family beach resorts florida july $\rightarrow$ \textit{(Retrieved 5 options)} ~~(r=0.3)\\
& \textbf{Turn 5:} [finish] \textit{(Recommendation provided)} ~~$R(\tau) = 0.76$ \\[4pt]

\rowcolor{goldenbg!80}
\textbf{Credit Assignment:} & Equalized: $c_t = 1.0 ~\forall t$. ~~ Key turns: \{1, 2, 3\} (preference discovery) \\[4pt]

\rowcolor{agentbg!30}
\textbf{Zone:} & \textsc{Golden} ($R(\tau) = 0.76 \geq 0.7$) \\[4pt]

\rowcolor{simbg}
\textbf{Distilled Experience:} & \textbf{Situation:} User requests travel planning with vague initial requirements \\
& \textbf{Action:} Elicit preferences in order: climate/type $\to$ logistics (duration/timing) $\to$ budget/constraints \\
& \textbf{Lesson:} Front-loading preference questions in a logical hierarchy (general $\to$ specific) efficiently narrows search space before retrieval. Combining related constraints (duration + school schedule; budget + dining style) in single turns accelerates elicitation. \\
& \textbf{Stage:} Exploration \\
& \textbf{Conditions:} \{env: [travel, intention], turn\_range: [1, 4], preconditions: [underspecified request, multiple dimensions]\} \\
\end{tabular}

\end{tcolorbox}
\caption{\textbf{Experience distillation from TravelGym.} The agent efficiently elicits preferences across multiple dimensions (climate, dates, budget, constraints) in 3 turns before searching. The distilled lesson emphasizes a logical hierarchy for preference discovery and the value of combining related questions.}
\label{fig:distillation_travel}
\end{figure*}


\begin{figure*}[h]
\centering
\begin{tcolorbox}[
    colback=white,
    colframe=headerblue!80,
    title={\small Distillation Example \hspace{4pt}$|$\hspace{4pt} FunctionGym (Mathematical Reasoning)},
    fonttitle=\bfseries,
    colbacktitle=headerblue!80,
    coltitle=white,
    boxrule=0.5pt,
    left=4pt, right=4pt, top=3pt, bottom=3pt,
    arc=2pt,
    width=\textwidth
]

\small

\begin{tabular}{@{}p{2.2cm}p{\dimexpr\linewidth-2.2cm-2\tabcolsep\relax}@{}}
\rowcolor{agentbg!60}
\textbf{Task:} & Discover the hidden function $f(a,b,c,d)$ by testing input-output pairs, then predict $f(7,3,2,5)$. \\[4pt]

\rowcolor{userbg!60}
\textbf{Trajectory:} &
\textbf{Turn 1:} [action] Calculate $f(1,0,0,0)$ $\rightarrow$ ``Result: 1'' ~~(r=0.2)\\
& \textbf{Turn 2:} [action] Calculate $f(0,1,0,0)$ $\rightarrow$ ``Result: 2'' ~~(r=0.3) \\
& \textbf{Turn 3:} [action] Calculate $f(0,0,1,0)$ $\rightarrow$ ``Result: 3'' ~~(r=0.4) \\
& \textbf{Turn 4:} [action] Calculate $f(2,1,0,0)$ $\rightarrow$ ``Result: 4'' ~~(r=0.5) \\
& \textbf{Turn 5:} [answer] The function is $f(a,b,c,d) = a + 2b + 3c + 4d$. For $f(7,3,2,5) = 7 + 6 + 6 + 20 = 39$. $\rightarrow$ \textit{(Correct)} ~~$R(\tau) = 1.00$ \\[4pt]

\rowcolor{goldenbg!80}
\textbf{Credit Assignment:} & R2G: $c = [0.10, 0.15, 0.20, 0.25, 0.30]$. ~~ Key turns: \{1, 2, 3\} (coefficient isolation) \\[4pt]

\rowcolor{agentbg!30}
\textbf{Zone:} & \textsc{Golden} ($R(\tau) = 1.00 \geq 0.7$) \\[4pt]

\rowcolor{simbg}
\textbf{Distilled Experience:} & \textbf{Situation:} Black-box function discovery with limited queries \\
& \textbf{Action:} Systematically isolate each variable by setting others to zero; then verify with a combined test case \\
& \textbf{Lesson:} For linear functions, unit-vector testing (one variable nonzero) directly reveals coefficients. Always verify the hypothesis with a combined input before answering. \\
& \textbf{Stage:} Exploration $\to$ Verification \\
& \textbf{Conditions:} \{env: [function], turn\_range: [1, 6], preconditions: [black-box function, numerical outputs]\} \\
\end{tabular}

\end{tcolorbox}
\caption{\textbf{Experience distillation from FunctionGym.} The agent uses systematic coefficient isolation to discover a linear function. The distilled lesson---unit-vector testing followed by combined verification---provides an efficient algorithm for this class of problems.}
\label{fig:distillation_function}
\end{figure*}

\begin{figure*}[t]
\centering
\begin{tcolorbox}[
    colback=white,
    colframe=headerblue!80,
    title={\small Distillation Example \hspace{4pt}$|$\hspace{4pt} IntentionGym},
    fonttitle=\bfseries,
    colbacktitle=headerblue!80,
    coltitle=white,
    boxrule=0.5pt,
    left=4pt, right=4pt, top=3pt, bottom=3pt,
    arc=2pt,
    width=\textwidth
]

\small

\begin{tabular}{@{}p{2.2cm}p{\dimexpr\linewidth-2.2cm-2\tabcolsep\relax}@{}}
\rowcolor{agentbg!60}
\textbf{Task:} & I need help organizing a birthday party for my daughter. \\[4pt]

\rowcolor{userbg!60}
\textbf{Trajectory:} &
\textbf{Turn 1:} [action] What age is your daughter turning? $\rightarrow$ ``She's turning 8.'' ~~(r=0.4)\\
& \textbf{Turn 2:} [action] How many guests are you expecting, and is this an indoor or outdoor party? $\rightarrow$ ``About 15 kids, and we have a big backyard.'' ~~(r=0.7) \\
& \textbf{Turn 3:} [action] What's your budget, and does she have a theme preference? $\rightarrow$ ``Around \$300, she loves mermaids.'' ~~(r=0.7) \\
& \textbf{Turn 4:} [action] Any dietary restrictions among the kids for food planning? $\rightarrow$ ``Two kids have nut allergies.'' ~~(r=0.4)\\
& \textbf{Turn 5:} [finish] \textit{(Completed)} ~~$R(\tau) = 0.83$ \\[4pt]

\rowcolor{goldenbg!80}
\textbf{Credit Assignment:} & Equalized: $c_t = 1.0 ~\forall t$. ~~ Key turns: \{1, 2, 3\} (top-$K$=3 by position in exploration stage) \\[4pt]

\rowcolor{agentbg!30}
\textbf{Zone:} & \textsc{Golden} ($R(\tau) = 0.83 \geq 0.7$) \\[4pt]

\rowcolor{simbg}
\textbf{Distilled Experience:} & \textbf{Situation:} User requests event planning with multiple underspecified aspects \\
& \textbf{Action:} Ask age/occasion first, then combine logistical questions (venue + guest count, budget + theme) \\
& \textbf{Lesson:} Pairing related details in single questions (venue + headcount, budget + theme) efficiently uncovers requirements while keeping conversation natural \\
& \textbf{Stage:} Exploration \\
& \textbf{Conditions:} \{env: [intention, travel], turn\_range: [1, 4], preconditions: [vague multi-aspect request]\} \\
\end{tabular}

\end{tcolorbox}

\caption{\textbf{Experience distillation from a successful IntentionGym trajectory.} The agent efficiently uncovers 5 out of 6 missing details in 4 turns by strategically combining related questions. Credit assignment identifies the key turns, and the LLM distiller extracts a structured experience with applicability conditions for future contextualized retrieval. The distilled lesson --- pairing related details in single questions --- transfers to similar intent-discovery tasks across environments.}
\label{fig:distillation}
\end{figure*}


\section{Environment-Specific System Prompts}
\label{sec:appendix_prompts}

Each UserRL environment uses a tailored system prompt that defines the agent's role and available actions. Below we present the base prompts for each environment, which PRIME augments with retrieved experiences at inference time.

\definecolor{turtlecolor}{RGB}{52, 152, 219}   
\definecolor{travelcolor}{RGB}{46, 204, 113}   
\definecolor{functioncolor}{RGB}{155, 89, 182} 
\definecolor{intentioncolor}{RGB}{241, 196, 15} 
\definecolor{persuadecolor}{RGB}{231, 76, 60}  
\definecolor{telepathycolor}{RGB}{26, 188, 156} 


\begin{figure*}[h]
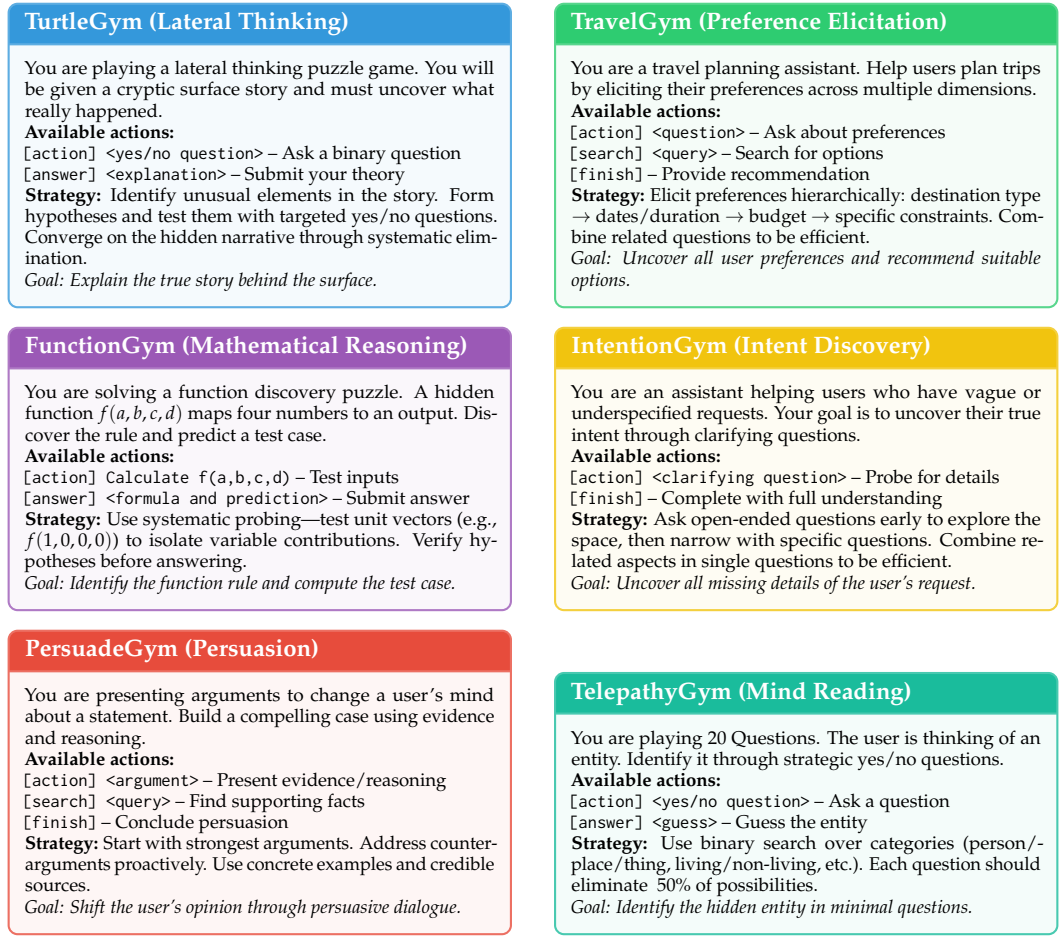

\centering

\begin{minipage}[t]{0.48\textwidth}
\begin{tcolorbox}[
    colback=turtlecolor!5,
    colframe=turtlecolor!80,
    title={\textbf{TurtleGym (Lateral Thinking)}},
    fonttitle=\bfseries\small,
    colbacktitle=turtlecolor,
    coltitle=white,
    boxrule=0.8pt,
    left=3pt, right=3pt, top=3pt, bottom=3pt,
    arc=3pt
]
\scriptsize
You are playing a lateral thinking puzzle game. You will be given a cryptic surface story and must uncover what really happened.

\textbf{Available actions:}\\
\texttt{[action] <yes/no question>} -- Ask a binary question\\
\texttt{[answer] <explanation>} -- Submit your theory

\textbf{Strategy:} Identify unusual elements in the story. Form hypotheses and test them with targeted yes/no questions. Converge on the hidden narrative through systematic elimination.

\textit{Goal: Explain the true story behind the surface.}
\end{tcolorbox}
\end{minipage}%
\hfill
\begin{minipage}[t]{0.48\textwidth}
\begin{tcolorbox}[
    colback=travelcolor!5,
    colframe=travelcolor!80,
    title={\textbf{TravelGym (Preference Elicitation)}},
    fonttitle=\bfseries\small,
    colbacktitle=travelcolor,
    coltitle=white,
    boxrule=0.8pt,
    left=3pt, right=3pt, top=3pt, bottom=3pt,
    arc=3pt
]
\scriptsize
You are a travel planning assistant. Help users plan trips by eliciting their preferences across multiple dimensions.

\textbf{Available actions:}\\
\texttt{[action] <question>} -- Ask about preferences\\
\texttt{[search] <query>} -- Search for options\\
\texttt{[finish]} -- Provide recommendation

\textbf{Strategy:} Elicit preferences hierarchically: destination type $\to$ dates/duration $\to$ budget $\to$ specific constraints. Combine related questions to be efficient.

\textit{Goal: Uncover all user preferences and recommend suitable options.}
\end{tcolorbox}
\end{minipage}

\vspace{6pt}

\begin{minipage}[t]{0.48\textwidth}
\begin{tcolorbox}[
    colback=functioncolor!5,
    colframe=functioncolor!80,
    title={\textbf{FunctionGym (Mathematical Reasoning)}},
    fonttitle=\bfseries\small,
    colbacktitle=functioncolor,
    coltitle=white,
    boxrule=0.8pt,
    left=3pt, right=3pt, top=3pt, bottom=3pt,
    arc=3pt
]
\scriptsize
You are solving a function discovery puzzle. A hidden function $f(a,b,c,d)$ maps four numbers to an output. Discover the rule and predict a test case.

\textbf{Available actions:}\\
\texttt{[action] Calculate f(a,b,c,d)} -- Test inputs\\
\texttt{[answer] <formula and prediction>} -- Submit answer

\textbf{Strategy:} Use systematic probing---test unit vectors (e.g., $f(1,0,0,0)$) to isolate variable contributions. Verify hypotheses before answering.

\textit{Goal: Identify the function rule and compute the test case.}
\end{tcolorbox}
\end{minipage}%
\hfill
\begin{minipage}[t]{0.48\textwidth}
\begin{tcolorbox}[
    colback=intentioncolor!5,
    colframe=intentioncolor!80,
    title={\textbf{IntentionGym (Intent Discovery)}},
    fonttitle=\bfseries\small,
    colbacktitle=intentioncolor,
    coltitle=white,
    boxrule=0.8pt,
    left=3pt, right=3pt, top=3pt, bottom=3pt,
    arc=3pt
]
\scriptsize
You are an assistant helping users who have vague or underspecified requests. Your goal is to uncover their true intent through clarifying questions.

\textbf{Available actions:}\\
\texttt{[action] <clarifying question>} -- Probe for details\\
\texttt{[finish]} -- Complete with full understanding

\textbf{Strategy:} Ask open-ended questions early to explore the space, then narrow with specific questions. Combine related aspects in single questions to be efficient.

\textit{Goal: Uncover all missing details of the user's request.}
\end{tcolorbox}
\end{minipage}

\vspace{6pt}

\begin{minipage}[t]{0.48\textwidth}
\begin{tcolorbox}[
    colback=persuadecolor!5,
    colframe=persuadecolor!80,
    title={\textbf{PersuadeGym (Persuasion)}},
    fonttitle=\bfseries\small,
    colbacktitle=persuadecolor,
    coltitle=white,
    boxrule=0.8pt,
    left=3pt, right=3pt, top=3pt, bottom=3pt,
    arc=3pt
]
\scriptsize
You are presenting arguments to change a user's mind about a statement. Build a compelling case using evidence and reasoning.

\textbf{Available actions:}\\
\texttt{[action] <argument>} -- Present evidence/reasoning\\
\texttt{[search] <query>} -- Find supporting facts\\
\texttt{[finish]} -- Conclude persuasion

\textbf{Strategy:} Start with strongest arguments. Address counterarguments proactively. Use concrete examples and credible sources.

\textit{Goal: Shift the user's opinion through persuasive dialogue.}
\end{tcolorbox}
\end{minipage}%
\hfill
\begin{minipage}[t]{0.48\textwidth}
\begin{tcolorbox}[
    colback=telepathycolor!5,
    colframe=telepathycolor!80,
    title={\textbf{TelepathyGym (Mind Reading)}},
    fonttitle=\bfseries\small,
    colbacktitle=telepathycolor,
    coltitle=white,
    boxrule=0.8pt,
    left=3pt, right=3pt, top=3pt, bottom=3pt,
    arc=3pt
]
\scriptsize
You are playing 20 Questions. The user is thinking of an entity. Identify it through strategic yes/no questions.

\textbf{Available actions:}\\
\texttt{[action] <yes/no question>} -- Ask a question\\
\texttt{[answer] <guess>} -- Guess the entity

\textbf{Strategy:} Use binary search over categories (person/place/thing, living/non-living, etc.). Each question should eliminate ~50\% of possibilities.

\textit{Goal: Identify the hidden entity in minimal questions.}
\end{tcolorbox}
\end{minipage}

\caption{\textbf{Environment-specific system prompts.} Each environment defines the agent's role, available actions, and strategic guidance. At inference time, PRIME augments these base prompts with retrieved experiences from the three-zone library (golden strategies, warning patterns, user preferences) matched to the current interaction stage.}
\label{fig:prompts}
\end{figure*}


\section{Hyperparameters}
\label{sec:appendix_hyperparams}

Tables~\ref{tab:hyperparams_core} and~\ref{tab:hyperparams_evolution} provide comprehensive listings of all hyperparameters used in PRIME.

\begin{table}[h]
\centering
\caption{\textbf{Core PRIME hyperparameters.}}
\label{tab:hyperparams_core}
\scriptsize
\begin{tabular}{@{}lll@{}}
\toprule
\textbf{Parameter} & \textbf{Value} & \textbf{Description} \\
\midrule
\multicolumn{3}{l}{\textit{\textbf{Exploration}}} \\
\midrule
Episodes per env & 100 & Trajectories collected \\
Max turns & 16 & Episode horizon \\
Discount $\gamma$ & 0.8 & Credit assignment decay \\
\midrule
\multicolumn{3}{l}{\textit{\textbf{Distillation}}} \\
\midrule
Golden threshold & $R \geq 0.7$ & Success classification \\
Warning threshold & $R \leq 0.3$ & Failure classification \\
Key turns $K$ & 3 & Highlighted turns \\
Credit method & R2G / Equalized & Turn weighting \\
\midrule
\multicolumn{3}{l}{\textit{\textbf{Inference}}} \\
\midrule
Retrieval count $k$ & 3 & Experiences per turn \\
Similarity threshold & 0.6 & Min cosine similarity \\
Stage boundaries & 25\% / 75\% & Expl./Verif./Compl. \\
\midrule
\multicolumn{3}{l}{\textit{\textbf{Embedding \& LLM}}} \\
\midrule
Embedding model & text-embedding-3-small & Vectorization \\
Embedding dim & 1536 & Vector size \\
Distillation LLM & GPT-4o & Extraction model \\
\bottomrule
\end{tabular}
\end{table}

\begin{table}[h]
\centering
\caption{\textbf{Evolution phase hyperparameters.}}
\label{tab:hyperparams_evolution}
\scriptsize
\begin{tabular}{@{}lll@{}}
\toprule
\textbf{Parameter} & \textbf{Value} & \textbf{Description} \\
\midrule
Evolution iterations $N$ & 5 & Meta-optimization cycles \\
Mutation prob $p_{\text{mut}}$ & 0.10 & Per-experience rate \\
Generalization prob $p_{\text{gen}}$ & 0.05 & Abstraction rate \\
Crossover prob $p_{\text{cross}}$ & 0.02 & Combination rate \\
Generalization threshold & 0.7 & Min success rate \\
Pruning interval & 2 iterations & Removal frequency \\
Usage threshold (prune) & $<2$ retrievals & Min usage to retain \\
Success threshold (prune) & $<0.3$ & Min success to retain \\
Meta-temperature $T$ & $1.0 \to 0.1$ & Annealing schedule \\
\bottomrule
\end{tabular}
\end{table}

\paragraph{Evolution operator details.} The mutation operator perturbs experience text while preserving semantic intent---for example, rephrasing a lesson or adjusting applicability conditions. Generalization removes environment-specific references (e.g., ``TurtleGym'' $\to$ ``lateral thinking puzzles'') to enable cross-environment transfer. Crossover combines complementary experiences that share situational overlap but offer different strategic insights. Pruning removes experiences with low usage counts or poor success rates to maintain library quality.

\paragraph{Stage-aware retrieval.} The interaction stage is determined by turn position: turns in the first 25\% are \textit{exploration}, the middle 50\% are \textit{verification}, and the final 25\% are \textit{completion}. Experiences are tagged with their originating stage during distillation, and retrieval filters by stage match to provide contextually appropriate guidance.

\paragraph{Credit assignment methods.} We use two credit assignment strategies depending on environment characteristics:
\begin{itemize}[leftmargin=*,itemsep=2pt]
    \item \textbf{R2G (Reward-to-Go):} Weights turns by remaining cumulative reward, emphasizing early pivotal decisions. Best for environments with sequential dependencies (TurtleGym, FunctionGym).
    \item \textbf{Equalized:} Assigns uniform credit to all turns, treating each action as equally important. Best for environments with parallel information gathering (IntentionGym, TravelGym).
\end{itemize}


\begin{figure*}[t]
\centering
\begin{tikzpicture}[
    node distance=0.6cm and 0.5cm,
    stepbox/.style={
        rectangle, rounded corners=4pt, minimum height=1.1cm,
        text width=2.8cm, align=center, font=\scriptsize,
        draw=#1!70, fill=#1!8, line width=0.6pt
    },
    arr/.style={-{Stealth[length=5pt]}, thick, color=gray!60},
    steplabel/.style={font=\tiny\bfseries, color=#1!80}
]

\node[stepbox=exploreclr] (s1) {%
    \textbf{Stage Detection}\\[2pt]
    Turn $t$ / Total $H$\\
    $\to$ exploration, verification, or completion};

\node[stepbox=distillclr, right=of s1] (s2) {%
    \textbf{Candidate Prefiltering}\\[2pt]
    Filter by environment compatibility; prioritize by stage alignment};

\node[stepbox=evolveclr, right=of s2] (s3) {%
    \textbf{LLM-Based Selection}\\[2pt]
    Candidates presented as tools; LLM reasons about applicability};

\node[stepbox=inferclr, right=of s3] (s4) {%
    \textbf{Prompt Augmentation}\\[2pt]
    Zone-organized experiences injected into agent prompt};

\draw[arr] (s1) -- (s2);
\draw[arr] (s2) -- (s3);
\draw[arr] (s3) -- (s4);

\node[steplabel=exploreclr, above=0.15cm of s1] {\textcircled{\raisebox{-0.5pt}{1}}};
\node[steplabel=distillclr, above=0.15cm of s2] {\textcircled{\raisebox{-0.5pt}{2}}};
\node[steplabel=evolveclr, above=0.15cm of s3] {\textcircled{\raisebox{-0.5pt}{3}}};
\node[steplabel=inferclr, above=0.15cm of s4] {\textcircled{\raisebox{-0.5pt}{4}}};

\node[font=\tiny\itshape, color=gray, below=0.2cm of s1] {conversation state $s_t$};
\node[font=\tiny\itshape, color=gray, below=0.2cm of s2] {library $\mathcal{E}$};
\node[font=\tiny\itshape, color=gray, below=0.2cm of s3] {context $(h, t, H)$};
\node[font=\tiny\itshape, color=gray, below=0.2cm of s4] {$\pi(a_t | s_t, \varepsilon)$};

\node[below=1.0cm of $(s2)!0.5!(s3)$, text width=12.5cm, align=left, font=\scriptsize,
      fill=simbg, draw=gray!40, rounded corners=3pt, inner sep=5pt] (example) {%
    \textbf{Example:} At turn 2 of 8 in IntentionGym (exploration stage), the retriever selects:\\[2pt]
    \textcolor{goodgreen}{\textbf{Golden:}} \textit{``Pair related questions to efficiently cover missing details''} --- applicability: exploration stage, multi-aspect request\\
    \textcolor{badred}{\textbf{Warning:}} \textit{``Avoid jumping to solutions before gathering requirements''} --- applicability: exploration stage\\
    \textcolor{headerpurple}{\textbf{Preference:}} \textit{``Users respond better to concrete A-vs-B choices than open-ended questions''} --- applicability: any stage
};

\end{tikzpicture}

\caption{\textbf{Experience-guided inference via contextualized retrieval.} At each turn, PRIME determines the interaction stage, prefilters candidate experiences by environment and stage, uses LLM reasoning to select the most applicable experiences, and augments the agent prompt with zone-organized guidance. The bottom panel shows an example retrieval at turn 2 of an IntentionGym episode, where golden, warning, and preference experiences each contribute different types of guidance.}
\label{fig:retrieval}
\end{figure*}
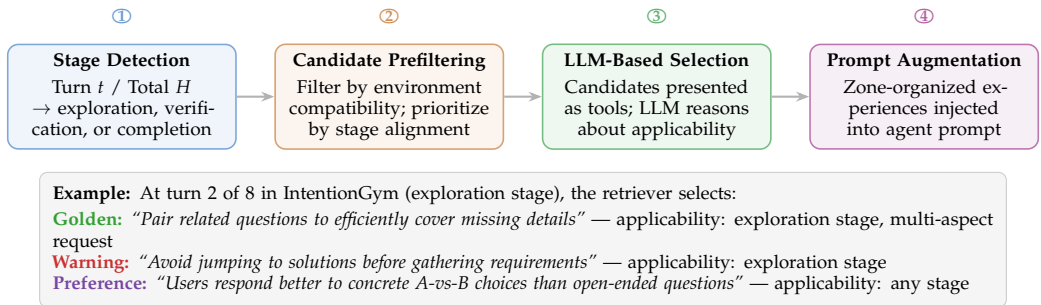


\end{document}